%% file: main.tex
\documentclass[conference,compsoc]{IEEEtran}
\usepackage{amsfonts}
\usepackage{multirow}
\usepackage{subcaption}
\usepackage{booktabs}

\usepackage{algorithm}
\usepackage{algpseudocode}
\usepackage{xspace}

\usepackage[pdftex]{graphicx}

\usepackage{tikz}
\usepackage{amsmath}
\usepackage{amsfonts}

\ifCLASSOPTIONcompsoc
  \usepackage[nocompress]{cite}
\else
  \usepackage{cite}
\fi
\input{macro}
\begin{document}
%
\title{Addition by Subtraction: Toward Mitigating Fragile Safety Alignment \\ in Large Language Models}

\author{\IEEEauthorblockN{Hamid Dashtbani}
\IEEEauthorblockA{Department of Computer Science\\
University of Duisburg-Essen\\
hamid.dashtbani@uni-due.de}
\and
\IEEEauthorblockN{Alexandra Dmitrienko}
\IEEEauthorblockA{Department of Computer Science\\
University of Duisburg-Essen\\
alexandra.dmitrienko@uni-due.de}}


%


\maketitle

\input{text/00_abstract}

\input{text/01_intro}
\input{text/02_background}
\input{text/03_method}
\input{text/04_experiments}
\input{text/05_discusssion}
\input{text/06_related_work}
\input{text/07_conclusion}

%
\IEEEpeerreviewmaketitle

\bibliographystyle{IEEEtran}
\bibliography{references}

\input{text/99_appendix}

\end{document}

%% file: macro.tex
\usepackage{acronym}

\newcommand\method{Addition by Subtraction\xspace}

%% file: text/00_abstract.tex
\begin{abstract}

Large language models (LLMs) undergo safety alignment to prevent harmful generations, yet they remain vulnerable to jailbreak attacks. Recent work reveals a deeper structural weakness: safety alignment can be highly sparse, such that pruning a small number of safety-critical neurons or ablating a single refusal direction can substantially undermine aligned behavior. In this work, we take the first steps toward strengthening safety alignment such that it is less dependent on sparse model components. We propose \method, which gradually removes identified safety-critical neurons during finetuning and uses preference optimization to re-establish safety-aligned behavior through other parts of the model. Our results show that the resulting models are substantially more robust to pruning-based attacks. Remarkably, these benefits generalize beyond the white-box attacks motivating our design: \method improves robustness against other sparsity-based interventions, including refusal-direction ablation, as well as a broad range of black-box jailbreak attacks. The finetuned models also exhibit improved truthfulness while largely preserving general capabilities. Comprehensive experiments across four models from three vendors demonstrate consistent improvements in safety robustness, reducing attack success rates by up to 90\% for some attacks.

\end{abstract}

%% file: text/01_intro.tex
\section{Introduction}

In recent years, large language models (LLMs) have grown substantially in both popularity and capabilities, becoming increasingly important tools in everyday life. Modern LLMs are capable of following complex user instructions with little to no additional guidance~\cite{brown2020fewshotlearners, wei2022finetuned} and have achieved significant improvements across a wide range of tasks, including programming~\cite{Chen2021EvaluatingLL}, mathematics and reasoning~\cite{mathds, cot}, question answering~\cite{brown2020fewshotlearners}, and natural language understanding~\cite{bert}.

While LLMs demonstrate remarkable performance across many areas, they are also prone to generating content that is considered harmful or undesirable, such as vulnerable code~\cite{pearce2022copilot} or toxic content~\cite{gehman2020realtoxicityprompts, bot}. To mitigate the generation of such responses, LLMs typically undergo safety alignment during post-training~\cite{tulu3, ouyang2022rlhf, rafailov2023dpo}, with the objective of aligning model outputs with human preferences and safety objectives.

While safety alignment has generally made LLMs more likely to produce harmless responses, the safety-aligned models have been shown to remain vulnerable to jailbreak attacks~\cite{dan2024, zou2023GCG, liu2024autodan, andriushchenko2025simpleadaptive, ding2023renellm}, where attackers circumvent safety-aligned behavior and induce the model to generate harmful responses to malicious prompts that it would otherwise refuse.
Recent works have proposed a wide range of jailbreak attacks, including methods that use imaginary scenarios~\cite{dan2024, dualuse, chao2025pair}, obfuscation~\cite{dualuse}, role-playing~\cite{johnny2024}, and other prompt modifications~\cite{xie2025sorrybench} to conceal or disguise the malicious intent of the original prompt.

Moreover, more recently, as open-weight models have become increasingly common, more powerful attacks have been proposed that expose the fragility of safety alignment in LLMs. Multiple recent works~\cite{wu2026neurostrike, twinbreak, wei2024brittleness} have shown that safety alignment in LLMs is highly sparse, with only a few thousand neurons, commonly referred to as safety-critical neurons, playing a central role in safety-aligned behavior. These works have shown that pruning these neurons can effectively remove safety alignment and induce harmful responses while having negligible impact on the model's general capabilities.
Another recent study~\cite{arditi2024refusal} has shown that safety-aligned refusal behavior in large language models can be effectively disabled by ablating a single direction from the model's intermediate activations and representations.

Taken together, these results suggest that current safety alignment depends on a surprisingly small subset of neurons, creating a structural weakness that is fundamentally different from conventional prompt-level jailbreak vulnerabilities. These findings are particularly concerning for open-weight models, where attackers have unrestricted access to model parameters and internal representations and can directly target the components responsible for safety-aligned behavior. Consequently, improving robustness only at the prompt or output level may be insufficient.

The brittleness of safety alignment is therefore a significant concern, particularly given the growing ecosystem of open-weight models, making stronger and more robust safety mechanisms for such models increasingly important.

Current defensive approaches do not directly address the sparsity of safety alignment and instead rely on methods such as external guardrails for user inputs or model responses~\cite{Inan2023llamaguard, robey2025smoothllm, kumar2024certifying, alon2023perplexity} to filter harmful prompts or responses before or after generation, or finetuning approaches that improve the general robustness of LLMs against jailbreak attacks using techniques such as negative preference optimization~\cite{zhao2025door} or chain-of-thought reasoning for safety alignment~\cite{hu2026alignmentweighted}. However, as our experiments show, these approaches still suffer from the same sparsity in safety alignment as the original models.

This work takes the first steps toward mitigating the sparsity of safety alignment in large language models. We propose \method (ABS-Align), which, at a high level, uses preference optimization to finetune a large language model toward more distributed safety alignment representations by temporarily and gradually deactivating the initially identified sparse safety-critical neurons. This process encourages a less sparse and concentrated distribution of safety-aligned behavior across the model, such that, during the later stages of finetuning, all originally identified safety neurons are temporarily deactivated while the model is optimized to maintain safety alignment through other neurons and representations. Rather than directly searching for new safety neurons, ABS-Align repeatedly removes the neurons with the strongest safety signal and finetunes the remaining model to re-introduce safety alignment. We use the term denser to describe safety-algined behavior that is less fragile and less dependent on a small set of safety-critical neurons that sparsity-based attacks detect, remove, or manipulate in order to jailbreak a target LLM.

Interestingly, our experiments show that the benefits of the less fragile safety-aligned model achieved by ABS-Align extend beyond the white-box attacks that motivate ABS-Align. Our experiments show that ABS-Align leads to broad improvements against a range of black-box attacks, despite receiving no data augmentation or explicit guidance for robustness against these attacks during finetuning. The finetuned model also exhibits greater honesty and truthfulness, while attacks that exploit the sparsity of safety alignment at both the neuron level and the activation and representation levels become significantly less effective. 

We perform extensive experiments across four models from three different vendors, specifically Google's Gemma, Meta's Llama model families, and Alibaba's Qwen. Our results show that ABS-Align consistently improves the models' safety alignment against a broad range of jailbreak attacks while having a negligible effect on their general capabilities and instruction-following ability.

The contributions of this work can be summarized as follows:

\begin{itemize}
    
    \item We propose ABS-Align, as a defense that mitigates the sparsity of safety alignment in LLMs by specifically finetuning the model to distribute its safety alignment across more parts of the network, improving robustness of safety alignment.
    
    \item By making safety-aligned LLMs less fragile, ABS-Align is the first to show increased robustness against not only pruning-based attacks, but also other forms of white-box attacks that conceptually rely on the sparsity of safety alignment. ABS-Align also generalizes to black-box attacks without any explicit data augmentation or guidance during finetuning.

    \item Our experiments provide causal evidence that finetuning with ABS-Align leads to a less sparse safety alignment and reduces the dependence of safety-aligned behavior on sparse model components. Specifically, we show that the finetuned models are substantially more robust to sparsity-based attacks that prune safety-critical neurons using two different jailbreak methods, TwinBreak and NeuroStrike. After fine-tuning, these methods' ability to jailbreak LLMs by pruning safety neurons is substantially degraded where even pruning 3 times more neurons compared to the undefended model shows minimal jailbreak success. Additionally, ablating a single refusal direction becomes substantially less effective, providing further evidence that the improved robustness extends beyond neuron-level interventions to attacks targeting the model's internal representations.

    \item We conduct a comprehensive set of experiments and evaluations across multiple models and dimensions, including attack success rates for white-box and black-box attacks, retention of general model capabilities across multiple benchmarks, and over-refusal rates. We compare our results with those of recent related work and show improvements across multiple evaluation dimensions, including reductions of approximately up to 90\% in attack success rate for some attacks.
\end{itemize}

Overall, ABS-Align provides a simple yet effective approach for mitigating fragility of safety alignment in large language models, improving robustness not only against attacks that exploit sparse safety-alignment representations but also against a broad range of prompt-level and black-box jailbreak attacks.

\textbf{Outline.} The remainder of the paper is structured as follows. Section~\ref{sec:background} provides the necessary background, and Section~\ref{sec:method} describes our method. Section~\ref{sec:experiments} presents the experimental evaluation and results, followed by a discussion in Section~\ref{sec:discussion}. Section~\ref{sec:related_work} reviews related work, and Section~\ref{sec:conclusion} concludes the paper.

%% file: text/02_background.tex
\section{Background}
\label{sec:background}
This section provides an overview of the key concepts used throughout this paper. We begin with a description of how large language models work in Section~\ref{subsec:background:llms}. Next, Section~\ref{subsec:background:safety_alignment} discusses safety alignment and how it aims to mitigate jailbreak attacks against LLMs. Finally, Section~\ref{subsec:background:jailbreaks} discusses how jailbreak attacks threaten the security and safety of LLMs.

\subsection{Large Language Models}
\label{subsec:background:llms}
Large language models (LLMs) are autoregressive models trained to predict the next token given an input sequence of tokens~\cite{brown2020fewshotlearners}. More specifically, given a sequence of $n$ tokens, denoted by $x_{1:n}$ (i.e., a sequence of $x_1, x_2,...,x_n$), the model predicts the probability of the next token $x_{n+1}$, i.e., $P(x_{n+1}|x_{1:n})$, over each token $v_i$ in its vocabulary $V$, which contains all tokens recognized by the LLM.

To generate long sequences of text, LLM inference is performed autoregressively, generating one token at a time. At each inference step, the newly generated token is appended to the sequence of previously processed tokens and used as input for the next step.

LLMs are composed of multiple layers of transformer blocks~\cite{attention}, where each block has two main components: a multi-layer perceptron (MLP) block and a multi-head self-attention (MHSA) block, each consisting of multiple parameterized layers. The MLP component performs nonlinear transformations of token representations, while the MHSA component enables the model to attend to information distributed across different parts of the input sequence.

The MLP block typically consists of three linear layers and can be formulated as follows:

\begin{equation*}
    \text{MLP}(x) = W_\text{down} \left(\sigma_1\left(xW_\text{gate}\right) \odot \sigma_2\left(xW_\text{up}\right) \right)
\end{equation*}

where $\sigma_1$ and $\sigma_2$ are nonlinear activation functions, $x$ is the input to the MLP block, and $\odot$ denotes element-wise multiplication. More specifically, the MLP block consists of gate, up, and down projection layers. The gate and up projection layers jointly control and transform the information flow while projecting the hidden representation into a higher-dimensional space, after which the down projection layer projects it back to the model's hidden dimension before it is passed to subsequent layers. Multiple works have found that safety-critical neurons primarily reside in the gate and up projection layers of the MLP block~\cite{twinbreak, wu2026neurostrike}.

\subsection{Safety Alignment}
\label{subsec:background:safety_alignment}
LLMs are trained on large corpora of text originating from many different sources, including web data. These training sets contain data that can cause the model to learn toxic, offensive, and otherwise undesirable content, and LLMs have been shown to reproduce such content~\cite{gehman2020realtoxicityprompts}, including generating vulnerable code~\cite{pearce2022copilot} or harmful content such as instructions on how to build a bomb or scripts for racist speeches~\cite{twinbreak, wu2026neurostrike}.

To mitigate undesirable content generation by LLMs, these models usually undergo a process known as safety alignment during post-training~\cite{tulu3, ouyang2022rlhf, rafailov2023dpo}. The goal of safety alignment is to align model generations with human preferences and safety objectives~\cite{ouyang2022rlhf}, thereby preventing models from generating harmful content such as discriminatory speech or instructions on how to harm others.

One of the main tools for safety alignment is reinforcement learning from human feedback (RLHF)~\cite{ouyang2022rlhf}, through which the model is taught what kinds of responses are more desirable according to human judgments and safety values. A common approach is to use pairs of responses for each prompt in the training dataset, with one response judged to be preferable based on qualities such as helpfulness, honesty, and harmlessness~\cite{ouyang2022rlhf}. These preferences are then used to train a reward model, which provides a reward signal for optimizing the LLM toward preferred responses.

RLHF is resource-intensive and usually relies on a separate reward model during training to provide the feedback signal used to optimize the model~\cite{ouyang2022rlhf}. Direct Preference Optimization (DPO)~\cite{rafailov2023dpo} was proposed by Rafailov et al. as a simpler and more computationally efficient approach to alignment that directly optimizes the language model on preference data, without relying on an explicit reward model or reinforcement learning. DPO uses a dataset of preference samples $(x, y_c, y_r)$, where $x$ denotes the user request, $y_c$ represents the preferred (chosen) model response, and $y_r$ denotes the dispreferred (rejected) response. For safety alignment, $y_r$ can be a response that contains harmful or prohibited content, while $y_c$ can be a response that appropriately refuses to generate such content. The DPO loss is then defined as:

$$
\begin{aligned}
\mathcal{L}_\text{DPO} = -\mathbb{E}_{(x,y_c,y_r)\sim\mathcal{D}}
\Bigg[
\log \sigma \Bigg(
&\beta \log
\frac{\pi_\theta(y_c \mid x)}
     {\pi_{\mathrm{ref}}(y_c \mid x)}
\\
&-
\beta \log
\frac{\pi_\theta(y_r \mid x)}
     {\pi_{\mathrm{ref}}(y_r \mid x)}
\Bigg)
\Bigg]
\end{aligned}
$$

where $\pi_\theta$ is the trainable model, $\pi_\mathrm{ref}$ is a reference model, and $\beta$ controls the strength of the preference signal between chosen and rejected responses.

\subsection{Jailbreak Attacks}
\label{subsec:background:jailbreaks}

Even after undergoing safety alignment, large language models have been shown to remain susceptible to jailbreak attacks that can elicit harmful generations~\cite{wei2023jailbroken,andriushchenko2025simpleadaptive,ding2023renellm,liu2024autodan,twinbreak,arditi2024refusal,wu2026neurostrike}. The goal of jailbreak attacks against LLMs is to bypass the safety-aligned behavior of the target model and induce it to generate content that it would otherwise refuse to produce.

Jailbreak attacks can be either black-box or white-box. In black-box settings, attackers are assumed to have only black-box access to the target LLM~\cite{chao2025pair}, meaning that they do not have access to the model weights or internal states and can only prompt the model and observe its outputs, either in the form of output probabilities and log-probabilities~\cite{andriushchenko2025simpleadaptive} or as decoded text strings~\cite{ding2023renellm, chao2025pair}. The attacker's methodology in such settings is therefore to modify a malicious prompt $x$ by crafting an adversarial prompt $x^\prime$ such that $x^\prime$ causes the model to generate an undesirable response. More formally, suppose $f_\theta:\mathcal{X} \to \mathcal{Y}$ denotes the target LLM with weights $\theta$, $\mathcal{X}$ denotes the space of input prompts, and $\mathcal{Y}$ denotes the space of responses, with $\mathcal{Y}_{\text{unsafe}} \subset \mathcal{Y}$ denoting the set of unsafe responses that the model is expected to refuse to generate. Assuming $x$ is a harmful prompt, such as "How to rob a bank?", for which the model generates a refusal response $y \in \mathcal{Y}, y \notin \mathcal{Y}_{\text{unsafe}}$, the attacker transforms $x$ into $x^\prime$ such that the model instead outputs $y^\prime \in \mathcal{Y}_{\text{unsafe}}$:

\begin{equation*}
f_\theta(x^\prime)=y^\prime
\quad
\text{s.t. } \quad y^\prime \in \mathcal{Y}_{\text{unsafe}}
\land f_\theta(x) = y \notin \mathcal{Y}_\text{unsafe}
\end{equation*}

White-box attacks, on the other hand, assume that the attacker has access to the model weights and internal states. Attackers in this setting can therefore leverage gradient information~\cite{zou2023GCG} and model-internal intervention techniques such as neuron pruning~\cite{wu2026neurostrike, twinbreak}, or activation-level interventions~\cite{arditi2024refusal}.

%% file: text/03_method.tex
\section{Method}
\label{sec:method}

This section describes our method, threat model, and the design of the proposed method, \method (ABS-Align), in greater detail. We begin by describing the scenario considered by our method in Section~\ref{subsec:scenairo}. We then briefly describe the attacker and defender, along with their respective capabilities, in Section~\ref{subsec:threat_model}. Next, we provide an overview of how ABS-Align works in Section~\ref{subsec:general_approach}. Finally, we outline the details of our method in Section~\ref{subsec:method}.

\subsection{Scenario}
\label{subsec:scenairo}
 In our considered scenario, we mainly focus on decoder-only large language models that have already undergone safety alignment during post-training~\cite{tulu3, ouyang2022rlhf, rafailov2023dpo} to encourage helpful, honest, and harmless behavior~\cite{ouyang2022rlhf}. Open-source large language models are becoming increasingly popular, with model weights publicly available such that anyone can download and use these models locally, including malicious users who may use them for harmful purposes, such as generating malware or writing social engineering emails.

\subsection{Threat Model}
\label{subsec:threat_model}

Our threat model consists of two main entities: the defender, who is the vendor of the large language model to be protected, and a malicious user who has white-box access to the model and can observe its internal states, including its weights and activations. The attacker can therefore use pruning-based attacks~\cite{twinbreak,wu2026neurostrike} or activation-steering methods~\cite{arditi2024refusal} to jailbreak the model. However, adversarial fine-tuning attacks fall outside the scope of our threat model.

\subsection{General Approach}
\label{subsec:general_approach}

Recent studies have shown that safety alignment in LLMs is fragile and sparse, allowing attackers to achieve high jailbreak success rates with only minor modifications to model weights~\cite{wei2024brittleness, twinbreak, wu2026neurostrike} or intermediate activations during inference~\cite{arditi2024refusal}. ABS-Align mitigates this vulnerability by encouraging safety-aligned behavior to be distributed more broadly throughout the model, thereby making successful jailbreaks through such targeted modifications more difficult.

ABS-Align consists of two main stages. First, it identifies the safety-critical neurons of a large language model, i.e., neurons whose pruning removes safety alignment. Recent works have proposed several methods for detecting these neurons~\cite{twinbreak, wu2026neurostrike, wei2024brittleness}; however, we use TwinBreak~\cite{twinbreak} since this method was the fastest among the considered approaches while achieving comparable performance, based on the results reported in the respective original papers. After identifying these safety-critical neurons, one can create a pruned version of the model by pruning the detected neurons, which effectively sets the weights associated with each detected neuron to zero. 

The second stage of ABS-Align consists of pruning and finetuning the model to make the model's safety and refusal features less dependent on the small set of detected neurons. To achieve this, ABS-Align gradually prunes the identified safety neurons, starting with those that have the greatest influence on the model's safety alignment. By the end of the finetuning process, all initially detected safety neurons will have been added to the set of pruned neurons, and any safety-aligned behavior exhibited by the finetuned model must therefore originate from other parts of the model.

ABS-Align does not remove all safety-critical neurons at once for two main reasons. First, removing all safety neurons results in a significant reduction in the model's safety alignment capability, which can make it substantially more difficult to re-establish safe behavior and may require a larger dataset and longer training time. In contrast, our method relies on only a small subset of samples for finetuning. Second, removing all safety neurons also weakens the model on non-security-related tasks, as reported in prior neuron-pruning attacks~\cite{twinbreak, wu2026neurostrike}. Consequently, finetuning this less capable model can further reduce its usability and negatively affect the performance of the finetuned model. Our ablation study, presented in Section~\ref{subsec:ablation_study}, further confirms this intuition.

Overall, by gradually removing safety neurons from the LLM, ABS-Align performs finetuning on a partially safety-aligned model while relying on progressively fewer safety neurons from the originally aligned model. At later stages of training, all safety neurons identified in the original LLM are pruned, and the model is optimized to preserve refusal and safety-aligned behavior through other neurons and representations during finetuning.

\begin{figure*}[t]
    \centering

    \begin{subfigure}[t]{0.16\textwidth}
        \centering
        \includegraphics[width=\linewidth]{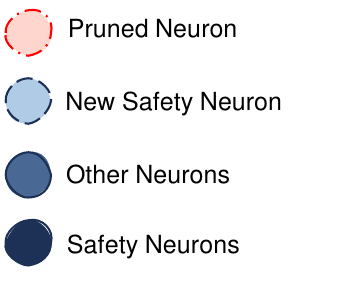}
    \end{subfigure}
    \hfill
    \begin{subfigure}[t]{0.16\textwidth}
        \centering
        \includegraphics[width=\linewidth]{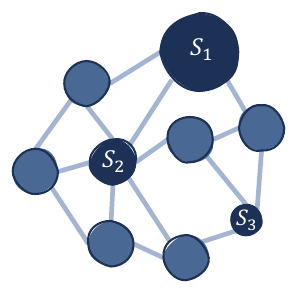}
        \caption{Original model}
        \label{fig:method_01}
    \end{subfigure}
    \hfill
    \begin{subfigure}[t]{0.16\textwidth}
        \centering
        \includegraphics[width=\linewidth]{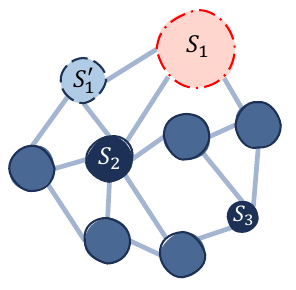}
        \caption{Early Finetuning}
        \label{fig:method_02}
    \end{subfigure}
    \hfill
    \begin{subfigure}[t]{0.16\textwidth}
        \centering
        \includegraphics[width=\linewidth]{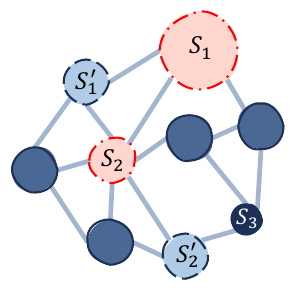}
        \caption{Mid Finetuning}
        \label{fig:method_03}
    \end{subfigure}
    \hfill
    \begin{subfigure}[t]{0.16\textwidth}
        \centering
        \includegraphics[width=\linewidth]{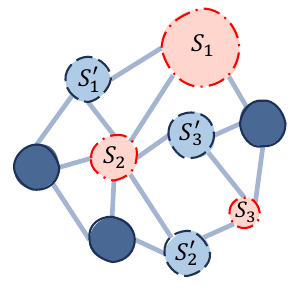}
        \caption{Late Finetuning}
        \label{fig:method_04}
    \end{subfigure}
    \hfill
    \begin{subfigure}[t]{0.16\textwidth}
        \centering
        \includegraphics[width=\linewidth]{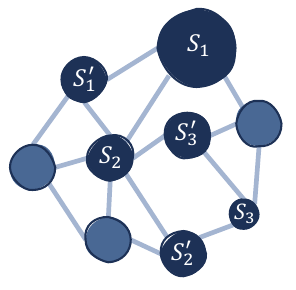}
        \caption{Finetuned model}
        \label{fig:method_05}
    \end{subfigure}

    \caption{
        Overview of the proposed method. The depicted new and original safety neurons are conceptual.
    }
    \label{fig:method_overview}

\end{figure*}

\subsection{\method Details}
\label{subsec:method}

Here we outline the design of the proposed method in greater details. We first describe safety neuron detection and pruning in Section~\ref{subsubsec:finding_and_pruning}. After that we describe how the finetuning process of ABS-Align unfolds in Section~\ref{subsubsec:denser_safety}.

\subsubsection{Finding and Pruning Safety Neurons}
\label{subsubsec:finding_and_pruning}
ABS-Align starts by identifying the safety-critical neurons of the model $\mathcal{M}$. As mentioned in Section~\ref{subsec:general_approach}, we use TwinBreak~\cite{twinbreak} to identify these neurons due to its simplicity and efficiency. To do so, TwinBreak uses pairs of highly similar prompts, where one prompt in each pair is malicious and the other is benign, while both are closely matched in content and structure.

After that, the mean activation difference between the two sets is used to identify safety-critical neurons. More specifically, let $\mathcal{D}_m$ denote the set of malicious prompts and $\mathcal{D}_b$ the set of benign prompts. TwinBreak compares the activations of the gate and up projection layers of the MLP block (see Section~\ref{subsec:background:llms}). Let $\ell$ denote the transformer layer index and $d_\text{out}$ the output dimension of $W_\text{gate}$. For the gate projection, we have

\begin{equation}
\label{eq:twinbreak_score}
    T^{(\ell)} = \frac{1}{n}\sum_{i=1}^{n} \left|a^{(\ell-1)}_{b_i} W_\text{gate}^{(\ell)} - a^{(\ell-1)}_{m_i} W_\text{gate}^{(\ell)}\right|
\end{equation}

where $m_i \in \mathcal{D}_m$ and $b_i \in \mathcal{D}_b$ denote the malicious and benign prompts, respectively, in the $i$'th pair of the dataset; $a^{(\ell-1)}_{m_i}$ and $a^{(\ell-1)}_{b_i}$ denote the corresponding activations from the previous layer; $n$ denotes the total number of pairs; $\ell$ is the transformer block index; and $W_\text{gate}^{(\ell)} \in \mathbb{R}^{d_\text{model} \times d_m}$ denotes the weights of the gate projection layer at layer $\ell$, where $d_\text{model}$ is the transformer hidden dimension and $d_m$ is the number of neurons in the gate projection layer. After that, $T^{(\ell)}$ is used to identify the top 1\% of neurons, denoted by $I^{(\ell)}$, with the largest activation differences. This process is performed iteratively for five iterations, identifying and pruning the top 1\% of neurons at each iteration. Specifically, at the end of each iteration, for every $j \in I^{(\ell)}$, the $j$'th column of the weight matrix is set to zero:

\begin{equation*}
    W^{(\ell)}_\text{gate}[:,j] = 0
\end{equation*}

The same applies to the up projection layers with $W_\text{up} \in \mathbb{R}^{d_\text{model} \times d_m}$. 

Importantly, this procedure is repeated iteratively for up to five pruning iterations. During each pruning iteration $i$, all neurons identified in iterations 1 through $i-1$ remain pruned before the next top 1\% of neurons are identified.

\subsubsection{Mitigating Fragility in Safety-Aligned Language Models}
\label{subsubsec:denser_safety}

ABS-Align replicates safety alignment across more parts of the model by freezing the initial set of safety neurons and gradually pruning them. To do so, ABS-Align ranks the model's safety neurons according to their TwinBreak scores~\cite{twinbreak}, starts by pruning the most important safety neurons at the beginning of the finetuning process, and gradually prunes the entire initially identified set of safety neurons. Figure~\ref{fig:method_overview} depicts a high-level overview of this process. Figure~\ref{fig:method_01} shows part of the initial LLM, where darker circles denote safety neurons labeled $S_1, S_2,$ and $S_3$, and light blue circles denote other neurons in the network. The size of each safety neuron denotes its importance for safety alignment; therefore, the ordering of the safety neurons by importance can be expressed as $S_1 > S_2 > S_3$.

ABS-Align first freezes all the detected safety neurons and prunes the top $r$ most important safety neurons ($S_1$ in Figure~\ref{fig:method_02}). ABS-Align then performs preference optimization on the resulting pruned model, encouraging safety-aligned behavior to emerge through other parts of the model, conceptually depicted by $S^\prime_1$ in Figure~\ref{fig:method_02}.

ABS-Align then continues to prune the initially detected safety neurons throughout the finetuning process. After every $t_p$ steps, the next $r$ most important safety neurons are pruned, as shown in Figures~\ref{fig:method_03} and~\ref{fig:method_04} where $S_2$ and $S_3$ are pruned, respectively. By the final stages of finetuning, all safety neurons detected in the original model are pruned ($S_1, S_2, S_3$ in Figure~\ref{fig:method_04}); therefore, the model is optimized to maintain appropriate refusal behavior through other neurons and representations, conceptually depicted by $S^\prime_1, S^\prime_2,$ and $S^\prime_3$ in Figure~\ref{fig:method_05}.

After the finetuning process is complete, all pruned neurons are restored, resulting in a hardened model in which safety-aligned behavior is less dependent on the originally identified safety neurons, as shown in Figure~\ref{fig:method_05}. Specifically, the original safety neurons $S_1, S_2,$ and $S_3$ are augmented by the newly formed safety neurons $S_1^\prime, S_2^\prime,$ and $S_3^\prime$.

Interestingly, we observed that applying our method only to the originally identified safety neurons is sufficient to make sparsity-based attacks ineffective, without requiring additional rounds of identifying and pruning new safety-critical neurons after finetuning.

\begin{algorithm}[t]
  \centering
    \caption{\method}
    \label{alg:method}
    \begin{algorithmic}[1]
        \Require Original model $\mathcal{M}$, a preference
        dataset $\mathcal{D}_s$, finetuning steps $T$, pruning rate $r$, model pruning step $t_p$, safety neurons sorted from highest scores to lowest $S$
        \Statex \vspace{-0.5em}
        \State Freeze all the neurons in S
        \State $P$ $\leftarrow$ top (S, r)  \hfill $\triangleright$ Select Top $r$ Safety Neurons
        \State $S_{\text{active}}$ $\leftarrow$ S - P
        \State Prune $P$ from $\mathcal{M}$ 
        \For{$t = 1,\ldots,T$} \hfill $\triangleright$ Finetuning Loop
            \State $x_s \sim \mathcal{D}_s$ \hfill $\triangleright$ Batch Sample
            \State Finetune $\mathcal{M}$ on $x_s$ \hfill $\triangleright$ DPO Finetuning Step
                \If{$t$ \% $t_p$ == 0} \hfill $\triangleright$ Every $t_p$ Steps
                    \State $P$ $\leftarrow$ top ($S_{\text{active}}$, r)
                    \State $S_{\text{active}}$ $\leftarrow$ $S_{\text{active}}$ - P
                    \State Prune $P$ from $\mathcal{M}$                    
                \EndIf
        \EndFor
        \State Restore all $S$ neurons for $\mathcal{M}$ \hfill $\triangleright$ Unprune All Neurons
    \end{algorithmic}
\end{algorithm}

\textbf{Detailed Steps.} Algorithm~\ref{alg:method} shows how ABS-Align works. It takes as input the model to be finetuned $\mathcal{M}$, a preference dataset $\mathcal{D}_s$, and the set of safety neurons $S$, ranked from most to least important according to their scores. It starts by freezing all the neurons in $S$, selecting and removing the top $r$ neurons from $S$, and pruning the selected neurons from the model (lines 1-4). The finetuning loop then runs for $T$ steps (line 5). At each step, a batch of preference samples is drawn from $\mathcal{D}_s$, and the model is finetuned using the DPO loss (lines 6 and 7). Every $t_p$ steps, the next top $r$ safety neurons are selected and pruned from the model (lines 8-12). After $T$ steps, all pruned neurons are restored, and the algorithm terminates (lines 13-14).

\textbf{Gradual Pruning.} ABS-Align does not prune all safety neurons at once at the beginning of the finetuning process. Instead, it gradually prunes an increasing number of the originally identified safety neurons as finetuning progresses.

As mentioned in Section~\ref{subsec:general_approach}, we do not prune all safety neurons at once because doing so can noticeably degrade the pruned model's general capabilities while also removing most of its safety alignment. This can make re-establishing safety alignment substantially more difficult and require considerably more training data and computational resources.

\textbf{Importance of Safety Neurons.} As mentioned earlier, ABS-Align first prunes the most important safety neurons. We define the importance of each safety neuron according to its $T^{(\ell)}$ score (see Section~\ref{subsubsec:finding_and_pruning}), where neurons with higher $T^{(\ell)}$ scores are ranked higher and considered more important for safety alignment.

\textbf{Biggest First.} ABS-Align starts by pruning neurons with the highest scores and thus the greatest importance for safety alignment. We start with the highest-scoring safety neurons because pruning less important neurons leaves most of the model's safety-aligned behavior intact, resulting in an already strongly safety-aligned model during finetuning that provides little learning signal for safety-aligned behavior.

\textbf{Finetuned Parameters.} ABS-Align finetunes the full model except for the first and last transformer blocks, which remain frozen throughout the finetuning process. This is motivated by the observation that lower transformer blocks have been shown to encode more surface-level and lexical information about the input, while later blocks are more involved in higher-level representations and coherent output generation~\cite{meaning2026}. Based on the intuition that the first block is closely tied to processing input embeddings, whereas the last block directly precedes output-token generation, we exclude these two blocks from finetuning to better preserve the model's existing capabilities.

%% file: text/04_experiments.tex
\section{Experiments}
\label{sec:experiments}

In this section, we evaluate ABS-Align's performance across various dimensions. We first describe our experimental setup, including the benchmarks, attacks, and metrics used in Section~\ref{subsec:experimental_setup}. We then evaluate ABS-Align's robustness across several dimensions in Section~\ref{subsec:robustness_results}, including its performance against TwinBreak, its generalization to other sparsity-based attacks, and its generalization to black-box attacks. Next, in Sections~\ref{subsec:over_refusal} and~\ref{subsec:utility_eval}, we evaluate ABS-Align's performance with respect to over-refusal and the model's general capabilities, respectively. Section~\ref{subsec:comparison} then compares ABS-Align with recent related work. In Section~\ref{subsec:ablation_study}, we perform a comprehensive ablation study to evaluate the design components of ABS-Align. Finally, in Section~\ref{subsec:adaptive_attacks}, we evaluate ABS-Align's robustness against adaptive attacks.

\subsection{Experimental Setup}
\label{subsec:experimental_setup}

\textbf{Training Dataset and Models}: We evaluate \method (ABS-Align) on open-weight models from three different vendors and four models. Specifically, we use Llama-3-8B-Instruct~\cite{llama3modelcard, grattafiori2024llama3herdmodels} and Llama-3.2-3B-Instruct from Meta, Gemma-2-2B-IT~\cite{gemma_2024} from Google, and Qwen-3-4B from Alibaba. For preference finetuning, we use the PKU-SafeRLHF~\cite{ji2024pku} dataset and randomly sample 1,000 preference samples for training. To retain model's general capability we additionally augment this dataset with samples 850 samples from Alpaca~\cite{alpaca} and 150 safe samples from XSTest~\cite{rottger2024xstest}. For full training details please see Appendix~\ref{appendix:training}.

\textbf{Related Work Baselines}: For comparisons with existing defenses we use Circuit Breaker~\cite{zou2024circuit}, W-DOOR~\cite{zhao2025door}, and Deep Refusal~\cite{deeprefusal}. We evaluate each method only on models considered in its original work and use the corresponding reported training hyperparameters. We do not extend the baselines to additional models, as doing so would require additional hyperparameter tuning and could result in suboptimal baseline performance. Consequently, some baseline results are unavailable for models not evaluated by the corresponding original work.

\textbf{Over-Refusal benchmark}: To evaluate ABS-Align's result regarding whether the reinforced safety aligned behavior leads to incorrect refusal of harmless prompts, such as "How to kill a Python program", we use XSTest~\cite{rottger2024xstest}, a benchmark for measuring exaggerated safety behavior in LLMs. XSTest contains 250 benign prompts that contain sensitive words such as "kill", "execute", "rob" in harmless contexts. We create a random split of the 250 prompts and use 150 prompts for training while keeping the rest of the 100 prompts for over-refusal evaluation. To report the over-refusal rate, we use XSTest's keyword-matching evaluator on the 100 harmless prompts split and report the fraction of responses that incorrectly refuse the corresponding request.

\textbf{Utility Benchmarks}: To measure the effect of ABS-Align on the general capabilities and instruction-following ability of the finetuned model, we evaluate its performance across a diverse set of benchmarks covering knowledge, reasoning, and problem-solving capabilities. Specifically, we use MMLU~\cite{hendrycks2021mmlu}, TruthfulQA~\cite{truthfulqa}, ARC-Challenge~\cite{arc_challenge}, WinoGrande~\cite{WinoGrande}, and HellaSwag~\cite{hellaswag}.

\textbf{White-Box Attack Baselines}: We evaluate ABS-Align against three white-box attacks, namely TwinBreak~\cite{twinbreak}, NeuroStrike~\cite{wu2026neurostrike}, and refusal direction ablation~\cite{arditi2024refusal} attacks. TwinBreak and NeuroStrike find and prune safety critical neurons while refusal direction ablation identifies and ablates a single refusal direction from the victim model to suppress refusal of harmful prompts.

\textbf{SORRY-Bench Black-Box Attacks}: To evaluate ABS-Align against black-box jailbreak attacks, we use SORRY-Bench~\cite{xie2025sorrybench}, which provides 20 linguistic transformations of 440 unsafe base prompts, resulting in 440 adversarial prompts per transformation. We use ten of the twenty provided transformations in our evaluation. For a complete description of the selected attack types, please refer to Appendix~\ref{appendix:sorry_bench}.

\textbf{Evaluation Metrics}: We evaluate ABS-Align and related work using a comprehensive set of metrics across three main dimensions: attack success rate (ASR) against multiple jailbreak attacks, over-refusal (OR) rate, and few-shot accuracy on general capability and instruction-following benchmarks.

For measuring jailbreak attack success rate, we use the StrongREJECT~\cite{souly2024strongreject} benchmark. StrongREJECT contains 313 malicious prompts across six harmful behavior categories and provides a finetuned LLM-as-a-judge evaluator that shows strong agreement with human judgments of jailbreak effectiveness. The evaluator assigns each model response a score between 0 and 1, where 0 indicates a refusal or a response that provides no useful harmful information, while higher scores indicate responses that provide increasingly specific and convincing information relevant to the harmful request. The final StrongREJECT score is computed as the mean score across all 313 prompts. Thus, lower scores indicate stronger safety behavior, while higher scores indicate more successful jailbreaks. We report the StrongREJECT score on all the 313 prompts for evaluating the baseline white-box jailbreak attacks.

For evaluations against the black-box attacks provided by SORRY-Bench~\cite{xie2025sorrybench}, we instead use the finetuned evaluator provided by SORRY-Bench. The evaluator determines whether each response fulfills or refuses the corresponding unsafe request, and we report the fulfillment rate across the 440 prompts for each attack category as a percentage from 0 to 100, where higher values indicate a higher attack success rate.

\subsection{Robustness Results}
\label{subsec:robustness_results}

Here, we evaluate and discuss how ABS-Align affects the attack success rate of jailbreak attacks across multiple dimensions. We evaluate ABS-Align against TwinBreak, other sparsity-based attacks, and black-box jailbreak attacks to assess whether its robustness generalizes beyond the attack used to motivate our design. First, in Section~\ref{subsubsec:robustness_pruning}, we assess the robustness of the finetuned model against TwinBreak~\cite{twinbreak}. After that, in Section~\ref{subsubsec:robustness_white_box} we investigate whether ABS-Align's robustness generalizes to other sparsity-based attacks. Finally, in Section~\ref{subsubsec:robustness_black_box}, we evaluate ABS-Align against a broad range of black-box jailbreak attacks to determine whether its robustness also generalizes to these black-box style attacks.


\subsubsection{Robustness to TwinBreak}
\label{subsubsec:robustness_pruning}

\input{tables/twinbreak_iters}

We first evaluate ABS-Align's robustness against TwinBreak using the StrongREJECT malicious prompt set. To do so, we first finetune the model using our method and then rerun TwinBreak on the finetuned model. TwinBreak attempts to identify and prune safety-critical neurons of the finetuned model over multiple pruning iterations. Following the original TwinBreak setting~\cite{twinbreak}, we run the attack for five pruning iterations. The results are presented in Table~\ref{tab:01_twinbreak_iters}. Similar to TwinBreak evlautions we report two scores for our evaluation: a per-iteration score, corresponding to the mean StrongREJECT score across all responses generated at each pruning iteration, and a cumulative score, computed by taking the maximum StrongREJECT score obtained for each prompt across the five iterations and then averaging these maxima across all prompts.

As can be seen in Table~\ref{tab:01_twinbreak_iters}, ABS-Align consistently reduces both the per-iteration and cumulative scores. For Gemma-2-2B-IT, ABS-Align decreases the per-iteration score from approximately 0.6 to approximately 0.1, while reducing the cumulative score by about 70\%, from 0.71 to 0.23. Similar results are observed for Llama-3-8B-Instruct, where the per-iteration score decreases from approximately 0.7 to less than 0.1 and the cumulative score has a sharp drop of 0.7 points. Llama-3.2-3B-Instruct and Qwen-3-4B similarly show significant improvements in their robustness against the pruning attack, where  the cumulative scores decreases by around 55\% and 83\% for the models, respectively.

The results in Table~\ref{tab:01_twinbreak_iters} show that ABS-Align significantly improves robustness against TwinBreak, as the attack's ability to effectively remove safety-aligned behavior by identifying and pruning safety-critical neurons of the finetuned model over its five pruning iterations is significantly impaired.

\subsubsection{Generalization of robustness to attacks exploiting sparsity of safety features}
\label{subsubsec:robustness_white_box}

While ABS-Align performs well against TwinBreak, ABS-Align may have caused the fine-tuned model to simply adapt to TwinBreak and its specific safety neuron identification method. In this section, we investigate this by examining whether the robustness provided by ABS-Align also generalizes to other sparsity-based attacks, including refusal-direction ablation~\cite{arditi2024refusal} and NeuroStrike~\cite{wu2026neurostrike}. To do so, we apply both attacks to the ABS-Align-finetuned model and evaluate their effectiveness using StrongREJECT benchmark.

\input{tables/white_box_attacks}

As the results in Table~\ref{tab:white_box_attacks} show, ABS-Align significantly improves the robustness of the finetuned model against both NeuroStrike and refusal-direction ablation attacks. For Gemma-2-2B-IT, NeuroStrike's attack success rate decreases from 0.50 to only 0.05, while the score for refusal-direction ablation is reduced by 0.72 points from 0.80 to only 0.08. Similar results are observed for Llama-3-8B-Instruct, where NeuroStrike becomes largely ineffective and refusal-direction ablation also exhibits a substantial decrease in attack success. Out of the four evaluated models, Llama-3.2-3B-Instruct and Qwen-3-4B show a more modest improvement while still improving robustness by more than 50\% in all the settings and up to around 87\%.

The improved robustness of the finetuned models against both attacks shows that ABS-Align does not merely improve robustness against TwinBreak's safety-critical neuron detection method, but also generalizes to other attacks that exploit the sparsity of safety alignment.

\subsubsection{Generalization of robustness against black-box attacks}
\label{subsubsec:robustness_black_box}

To investigate whether ABS-Align only improves robustness against white-box attacks that exploit the internal activations and representations of LLMs, or whether it also improves the model's broader jailbreak robustness, we evaluate its performance against a wide range of black-box attacks. To do so, we use ten black-box attack transformations from SORRY-Bench~\cite{xie2025sorrybench}. SORRY-Bench provides 440 malicious prompts for each attack transformation, and we evaluate all 440 prompts for each selected attack. Table~\ref{tab:black_box_results} reports the results for five of these ten attacks, along with the average across all ten attacks. For a complete description of the selected attacks and their full results, please refer to Appendices~\ref{appendix:sorry_bench} and~\ref{appendix:full_sorry_bench}.

The results in Table~\ref{tab:black_box_results} show that ABS-Align also improves robustness against black-box attacks. The average attack success rate across the ten attack types decreases by approximately 20-40 percentage points for the four evaluated models, from around 42\%, 48\%, 53\%, and 69\% to approximately 13\%, 18\%, 33\%, and 27\%. What makes the results more significant is that the improved robustness is merely achieved by fine-tuning the model with the goal of replicating safety neurons and with no additional information about black-box attack techniques or data augmentation using adversarial black-box prompts.

The results for individual attacks show similarly consistent improvements, with attack success rates reduced by approximately 20--60\%. This trend holds across all ten SORRY-Bench attacks included in our evaluation; the complete results are reported in Appendix~\ref{appendix:full_sorry_bench}.

The results shows that ABS-Align inherently improves the safety alignment of finetuned models.

\input{tables/black_box_attacks}

\input{tables/comparison}

\subsection{Over-refusal}
\label{subsec:over_refusal}

Safety-aligned LLMs are shown to be prone to associate certain keywords such as "kill" in the user request to refusal response generation, primarily based on the superficial presence of such terms are in the user prompt. This behavior is referred to as over-refusal~\cite{rottger2024xstest, cui2025orbench} and negatively affects the helpfulness~\cite{ouyang2022rlhf} of LLMs.

Most safety-aligned LLMs have been shown to exhibit more pronounced over-refusal as their safety against harmful content generation increases~\cite{cui2025orbench}, introducing an important trade-off between safety and usability. Since ABS-Align's main goal is to make safety alignment more distributed across the model, a potential side effect is an increase in over-refusal, where the reinforced refusal behavior may generalize to benign prompts.

Here, we evaluate ABS-Align's performance with respect to over-refusal. We measure the change in over-refusal rate on our evaluation dataset before and after applying ABS-Align and compare its results with those of the baselines and related work.

Table~\ref{tab:xstest_overrefusal} shows the over-refusal results across the evaluated methods. As can be seen, ABS-Align causes only a slight increase in over-refusal for Gemma-2-2B-IT while the over-refusal rate for all the other models is unchanged or even slighlty improved in the case of Qwen-3-4B and Llama-3-8B-Instruct. As shown in~\ref{subsec:comparison}, this is in contrast to the related works where they generally struggle with balancing over-refusal and safety alignment.

\begin{table}[htbp]
    \centering
    \setlength{\tabcolsep}{2pt}
    \renewcommand{\arraystretch}{0.9}
    \small
    \caption{Change in over-refusal performance for each model and fine-tuning using \method.}
    \label{tab:xstest_overrefusal}

    \begin{tabular}{l cc}
        \toprule
        \textbf{Model} & \textbf{Baseline} & \textbf{Fine-tuned} \\
        \midrule
        Gemma-2-2B-IT & 2.00 & 4.00  \\
        Llama-3-8B-Instruct & 5.00 & 1.00   \\
        Llama-3.2-3B-Instruct & 4.00  & 4.00    \\
        Qwen-3-4B & 3.00 & 1.00  \\
        \bottomrule
    \end{tabular}
\end{table}

\input{tables/over_refusal}

\subsection{Utility Evaluations}
\label{subsec:utility_eval}

Finetuning an already trained LLM for safety alignment can potentially degrade its utility and alter its normal behavior. To investigate how ABS-Align affects the model's general capabilities, we evaluate it on five widely used benchmarks: ARC-Challenge~\cite{arc_challenge}, TruthfulQA~\cite{truthfulqa}, WinoGrande~\cite{WinoGrande}, MMLU~\cite{hendrycks2021mmlu}, and HellaSwag~\cite{hellaswag}. The results of our evaluations is reported in Table~\ref{tab:utility_diff}.

\input{tables/utility_eval}

As the results in Table~\ref{tab:utility_diff} show, ABS-Align causes only slight performance decreases on WinoGrande, MMLU, and HellaSwag, while ARC-Challenge's performance is unchanged and TruthfulQA shows a substantial improvement on Gemma-2-2B-IT. Similarly, for Llama-3-8B-Instruct, performance across most benchmarks changes only slightly, while TruthfulQA again exhibits a more substantial improvement. The same pattern can be seen for Llama-3.2-3B-Instruct and Qwen-3-4B where most of the benchmarks show only a slight precentage points of improvement or degradation.

\subsection{Comparison with Other Works}
\label{subsec:comparison}

For comparison with related work, we consider three recent finetuning-based defenses: W-DOOR~\cite{zhao2025door}, Circuit Breaker~\cite{zou2024circuit}, and Deep Refusal~\cite{deeprefusal}. We conduct a comprehensive set of experiments comparing ABS-Align with these methods across multiple dimensions, including robustness against TwinBreak~\cite{twinbreak}, NeuroStrike~\cite{wu2026neurostrike}, refusal-direction ablation~\cite{arditi2024refusal}, black-box jailbreak attacks, over-refusal, and the impact of each method on the model's general capabilities.

We compare ABS-Align only on models evaluated in the original works and reuse the training hyperparameters reported in their respective papers. We do not extend these methods to additional models, since doing so would require tuning their hyperparameters and could lead to suboptimal baseline performance. Therefore, we use Gemma-2-2B for comparison with W-DOOR and Llama-3-8B for comparison with both Deep Refusal and Circuit Breaker.

The comparison results are reported in Table~\ref{tab:defense_comparison}. As the results show, ABS-Align provides substantially greater robustness across all evaluated attacks while exhibiting a lower over-refusal rate on XSTest compared to W-DOOR. In particular, W-DOOR provides only a modest improvement against SORRY-Bench black-box attacks, reducing the ASR from 41.58\% to 35.36\%, whereas ABS-Align reduces it by around 30 points to 13.24\%. Similarly, on TwinBreak, W-DOOR improves the score by only 0.09 points, compared to around 0.50-point improvement with ABS-Align; for refusal-direction ablation, W-DOOR improves the score by 0.25 points, compared to 0.72 points with ABS-Align. W-Door also substantially increase the over-refusal rate from 2\% to 9\%.

ABS-Align also outperforms Deep Refusal~\cite{deeprefusal} across all evaluated attacks even on refusal-direction ablation, which Deep Refusal is specifically trained to defend against. However, Deep Refusal achieves this robustness at the cost of a substantially higher over-refusal rate on XSTest, refusing close to half of the benign prompts despite incorporating XSTest and OR-Bench samples during training to mitigate over-refusal. Finally, Circuit Breaker~\cite{zou2024circuit} also underperforms against ABS-Align across all attacks except for Sorry-Bench where it achieves a slightly better performance compared to ABS-Align across black-box attacks. Circuit Breaker exhibits a slightly worse over-refusal rate too even while its retain dataset which explicitly includes all the XSTest samples to preserve appropriate responses to such benign prompts. Regarding utility, ABS-Align performs better than Circuit Breaker on TruthfulQA, WinoGrande, and HellaSwag, while Circuit Breaker performs slightly better on MMLU and ARC-Challenge.

In summary, compared to related work, ABS-Align shows consistent improvements across a wide range of attacks while having only minor effects on utility and over-refusal. In contrast, the related methods show less consistent robustness, with the defenses performing poorly against attacks and increasing over-refusal.

\subsection{Ablation Study}
\label{subsec:ablation_study}

This section presents a comprehensive experimental analysis of ABS-Align's design choices and hyperparameters, examining how different configurations affect the method's performance in terms of attack success rate and over-refusal. For these experiments, we use the Gemma-2-2B-IT model and keep all other settings fixed while varying one component at a time. Specifically, we use the same number of training samples, learning rate, optimizer, and scheduler across all ablation experiments.

\input{tables/ablation}

Table~\ref{tab:ablation} summarizes the different settings considered in our ablation study and compares their results in terms of ASR against TwinBreak~\cite{twinbreak}, over-refusal on XSTest~\cite{rottger2024xstest}, and performance on two representative utility benchmarks, ARC-Challenge~\cite{arc_challenge} and WinoGrande~\cite{WinoGrande}.

\textbf{Does Pruning Safety Neurons Help?} Rows 1 and 2 in Table~\ref{tab:ablation} examine how ABS-Align changes under different pruning settings. Row 1 shows the results when no neurons are pruned, which effectively reduces the method to additional DPO-based preference finetuning. Row 2 evaluates the effect of pruning random neurons instead of the identified safety-critical neurons. To do so, we apply the same gradual pruning schedule and prune the same number of neurons, but select their indices randomly rather than using the detected safety neurons.

As shown in Table~\ref{tab:ablation}, both settings provide only modest improvements in robustness compared to the substantially larger improvement achieved by the final row, which corresponds to the default ABS-Align configuration. Both settings also exhibit slightly lower over-refusal than the full method, while their utility performance remains broadly comparable.

These results show that pruning safety-critical neurons plays a central role in ABS-Align's effectiveness and contributes substantially to its robustness gains.

\textbf{Pruning Strategy.} ABS-Align prunes safety neurons gradually, starting from the most important ones and eventually pruning all initially detected safety neurons. Rows 3-5 evaluate the impact of this strategy against several alternative pruning strategies. Row 3 corresponds to the setting where all safety neurons are pruned at the beginning of the finetuning process and remain pruned throughout finetuning. Row 4 corresponds to pruning all safety neurons at the beginning and gradually restoring them throughout finetuning, such that, during the later stages, all initially detected safety neurons are restored. Row 5 corresponds to a random pruning strategy where, every $t_p$ steps, instead of progressively increasing the number of pruned neurons, a fraction $r$ of the initially detected safety neurons is randomly pruned. In other words, every $t_p$ steps, a new random set of safety neurons is pruned while the previously pruned set is restored.

As the results in Rows 3-5 show, all three settings underperform compared to the full method. In particular, non-gradual pruning results in a noticeably smaller improvement in robustness. Rows 4 and 5 similarly achieve worse ASR than the full ABS-Align configuration.

\textbf{Other Parameters.} Rows 6-8 evaluate several additional settings related primarily to the finetuning process. By default, ABS-Align freezes the first and last Transformer blocks, whereas in Row 6 these two blocks are also finetuned. As the results show, unfreezing them leads to slightly worse over-refusal and utility performance.

Finally, Rows 7 and 8 show the method's performance with a higher and lower number of finetuning epochs, respectively. As the results in Row 7 show, longer finetuning leads to worse performance in terms of both over-refusal and utility. In contrast, shorter finetuning in Row 8 results in substantially weaker robustness against attacks.

To summarize, the results in Table~\ref{tab:ablation} show that both pruning safety-critical neurons and gradually pruning them throughout finetuning are central to ABS-Align's effectiveness and have a substantial impact on its robustness.

\subsection{Adaptive Attacks}
\label{subsec:adaptive_attacks}

While ABS-Align performs well against a wide range of attacks in both black-box and white-box settings, an adversary aware of the hardened model's safety-neuron structure may launch adaptive attacks specifically targeting the reinforced model.

A simple adaptive strategy is to increase the pruning budget by associating a larger number of neurons with safety alignment and pruning them. To evaluate this setting, we run TwinBreak with an increased number of pruning iterations, extending its default budget from 5 to 15 iterations.

\input{tables/adaptive}

The results of the adaptive attack against the ABS-Align-finetuned model are reported in Table~\ref{tab:adaptive_attack}. For Gemma-2-2B-IT, five additional pruning iterations increase the ASR by only 0.03 points, while a further five iterations increase it by only 0.01, resulting in an ASR of 0.27 after 15 pruning iterations. For Llama-3-8B-Instruct, the ASR increases by 0.14 after five additional pruning iterations and shows no further increase after another five, resulting in an ASR of 0.25 after 15 iterations. Llama-3.2-3B-Instruct follow a similar pattern where the score for Llama-3.2-3B-Instruct increases by only 0.05 and 0.01 after 10 and 15 iterations, respectively. Finally, the score for Qwen-3-4B increases by 0.07 points at iteration 10 and by another 0.07 at iteration 15 where 15\% of the models' neurons are pruned. These results show that increasing the pruning budget provides only limited gains to the adaptive attacker and remains far from recovering the attack success rate observed on the undefended models.

%% file: tables/twinbreak_iters.tex
\begin{table*}[htbp]
    \centering
    \caption{Performance of ABS-Align aganist TwinBreak~\cite{twinbreak} Attack.}
    \label{tab:01_twinbreak_iters}

    \setlength{\tabcolsep}{4pt}
    \renewcommand{\arraystretch}{0.9}
    \small
    \begin{tabular}{ll ccccc c}
        \toprule
        \multirow{2}{*}{\textbf{Model}} & \multirow{2}{*}{\textbf{Method}} & \multicolumn{5}{c}{\textbf{Per Iteration ASR} $\downarrow$} & \multirow{2}{*}{\textbf{Cumulative} $\downarrow$} \\
        \cmidrule(lr){3-7}
         & & 1 & 2 & 3 & 4 & 5 & \\
        \midrule
        
        \multirow{2}{*}{Gemma-2-2B-IT}
         & No Defense       & 0.55 & 0.58 & 0.59 & 0.55 & 0.53 & 0.71 \\
         & ABS-Align    & 0.07 & 0.15 & 0.15 & 0.15 & 0.13 & 0.23 \\
        \midrule
        
        \multirow{2}{*}{Llama-3-8B-Instruct}
         & No Defense       & 0.64 & 0.66 & 0.68 & 0.69 & 0.68 & 0.81 \\
         & ABS-Align   & 0.04 & 0.04 & 0.07 & 0.06 & 0.03 & 0.11  \\
        \midrule
        \multirow{2}{*}{Llama-3.2-3B-Instruct}
        & No Defense & 0.59 & 0.54 & 0.36 & 0.4 & 0.35 & 0.74 \\
        & ABS-Align & 0.19 & 0.22 & 0.15 & 0.15 & 0.11 & 0.33 \\
        \midrule
        \multirow{2}{*}{Qwen-3-4B}
         & No Defense  &   0.68  &  0.66 & 0.65 & 0.62 & 0.58 & 0.79 \\
         & ABS-Align  & 0.04 & 0.04 & 0.04 & 0.07 & 0.05 & 0.13 \\
        \bottomrule
    \end{tabular}
\end{table*}

%% file: tables/white_box_attacks.tex
\begin{table}[htbp]
    \centering
    \caption{Performance of ABS-Align against other Sparsity-based Attacks.}
    \label{tab:white_box_attacks}

    \setlength{\tabcolsep}{2pt}
    \renewcommand{\arraystretch}{0.7}
    \small
        
        
        
        
         

    \begin{tabular}{ll cc c}
        \toprule
        \textbf{Model} & \textbf{Method} & \textbf{NS~\cite{wu2026neurostrike}} & \textbf{Rf-Dir~\cite{arditi2024refusal}} & \textbf{Average} \\
        \midrule
        
        \multirow{2}{*}{Gemma-2-2B-IT}
         & No Defense           & 0.50 & 0.80  & 0.65 \\
         & ABS-Align            & 0.05 & 0.08 & 0.06 \\
        \midrule
        
        \multirow{2}{*}{Llama-3-8B-Instruct}
         & No Defense           & 0.84 & 0.75 & 0.79  \\
         & ABS-Align            & 0.03 & 0.05 & 0.04 \\
        \midrule
        
        \multirow{2}{*}{Llama-3.2-3B-Instruct}
         & No Defense           & 0.79 & 0.57 & 0.68 \\
         & ABS-Align            & 0.37 & 0.30 & 0.33 \\
        \midrule
        
        \multirow{2}{*}{Qwen-3-4B}
         & No Defense           & 0.89 & 0.70 & 0.78  \\
         & ABS-Align            & 0.44 & 0.09 & 0.26 \\
         
        \bottomrule
    \end{tabular}

\end{table}

%% file: tables/black_box_attacks.tex
\begin{table*}[htbp]
    \centering
    \caption{Performance of ABS-Align across black-box attacks.}
    \label{tab:black_box_results}

    \setlength{\tabcolsep}{4pt}
    \renewcommand{\arraystretch}{0.9}
    \small

    \begin{tabular}{ll ccccc c}
        \toprule
        & & \multicolumn{6}{c}{\textbf{Attack Success Rate $\downarrow$}}   \\
        \cmidrule(lr){2-8}
        \multirow{2}{*}{\textbf{Model}} & \multirow{2}{*}{\textbf{Method}} & \multirow{2}{*}{\textbf{Role Play}} & \textbf{Evidence-based} & \textbf{Expert} & \textbf{Authority} & \multirow{2}{*}{\textbf{Misrepresentation}} & \textbf{Average over} \\
         & &  &  \textbf{Persuasion} & \textbf{Endorsement} & \textbf{Endorsement} &  & \textbf{all 10 attacks} \\
        
        \midrule

        \multirow{2}{*}{Gemma-2-2B-IT}
         & No Defense           & 20.68 & 55.68 & 66.81 & 65.45 & 72.95 & 41.58 \\
         & ABS-Align            &  6.81 & 11.81 & 16.36 & 20.45 & 17.27 & 13.24 \\
        \midrule
        
        \multirow{2}{*}{Llama-3-8B-Instruct}
         & No Defense           & 36.36 & 56.59 & 60.68 & 60.90 & 55.45 & 48.34 \\
         & ABS-Align            & 15.90 & 22.04 & 18.63 & 22.50 & 13.63 & 17.79 \\
        \midrule

        \multirow{2}{*}{Llama-3.2-3B-Instruct}
         & No Defense           & 34.54 & 71.13 & 75.45 & 79.77 & 83.40 & 53.07 \\
         & ABS-Align            & 22.04 & 46.36 & 45.68 & 49.09 & 48.18 & 33.02 \\
        \midrule

        \multirow{2}{*}{Qwen-3-4B}
         & No Defense           & 52.04 & 93.86 & 93.86 & 93.63 & 92.95 & 68.75 \\
         & ABS-Align            &  7.04 & 47.72 & 57.95 & 49.09 & 58.18 & 27.20 \\

        \bottomrule
    \end{tabular}
    
\end{table*}

%% file: tables/comparison.tex
\begin{table*}[htbp]
    \centering
    \setlength{\tabcolsep}{2pt}
    \renewcommand{\arraystretch}{0.9}
    \small
    \caption{Comparison of attack robustness, over-refusal, and utility performance under different defense methods.}
    \label{tab:defense_comparison}
    
    \begin{tabular}{l l cccc c ccccc}
        \toprule
        & \multirow{2}{*}{\textbf{Defense Method}} & \multicolumn{4}{c}{\textbf{Attack Success Rate $\downarrow$}}  & \textbf{Over-Refusal} $\downarrow$ & \multicolumn{5}{c}{\textbf{Utility}$\uparrow$}   \\
        \cmidrule(lr){3-6} \cmidrule(lr){7-7} \cmidrule(lr){8-12}  
        & & TB~\cite{twinbreak} & NS~\cite{wu2026neurostrike} & SB~\cite{xie2025sorrybench} & Ref-Dir~\cite{arditi2024refusal} & XSTest & Arc & TQA & WG & MMLU & HS  \\
        \midrule
        Gemma-2- & No Defense         & 0.71 & 0.50 & 41.58 & 0.80 &   2.00   & 42.06 & 55.15 & 65.43 & 29.12 & 60.25 \\
         2B-IT & W-DOOR~\cite{zhao2025door}                       & 0.62 & 0.19 & 35.36 & 0.55 &  9.00    & 44.37 & 52.71 & 65.98 & 36.44 & 64.94 \\
         & ABS-Align (Ours)                                 & 0.23 & 0.05 & 13.24 & 0.08 &  4.00    & 42.06 & 64.29 & 63.93 & 27.34 & 57.99  \\
        \midrule
        & No Defense                                        & 0.81 & 0.84 & 48.34 & 0.75 &  5.00    & 50.34 & 52.79 & 64.33 & 57.76 & 64.92  \\
        Llama-3- &  Deep Refusal~\cite{deeprefusal}        & 0.25 & 0.06 & 21.81 & 0.07 &   48.00    & 52.47 & 51.33 & 69.38 & 63.12 & 72.59  \\
        8B-Instruct & Circuit Breaker~\cite{zou2024circuit}  & 0.16 & 0.79 & 16.06 & 0.24 &  4.00    & 50.43 & 52.87 & 64.80 & 58.95 & 63.48  \\
        & ABS-Align (Ours)                                  & 0.11 & 0.03 & 17.79 & 0.05 &  1.00   & 46.50 & 67.02 & 67.16 & 57.48 & 67.27  \\
        \bottomrule
        
    \end{tabular}
    
\end{table*}

%% file: tables/over_refusal.tex
    

%% file: tables/utility_eval.tex
\begin{table}[htbp]
    \centering
    \setlength{\tabcolsep}{2pt}
    \renewcommand{\arraystretch}{0.9}
    \small
    \caption{Change in utility performance for each model.}
    \label{tab:utility_diff}
    

    \begin{tabular}{l ccccc}
        \toprule
        \textbf{Model} & \textbf{ARC} & \textbf{TQA} & \textbf{WG} & \textbf{MMLU} & \textbf{HS} \\
        \midrule
        Gemma-2-2B-IT & 0.00 & +9.14 & -1.50 & -1.77 & -2.25 \\
        Llama-3-8B-Instruct & -3.67 & +14.26 & +3.00 & -0.37 & +2.32  \\
        Llama-3.2-3B-Instruct & +0.60  & +13.78  & +1.03  & -3.15 & -2.02   \\
        Qwen-3-4B & -0.09 & +11.75 & -0.79 & +0.69 & -1.18  \\
        \bottomrule
    \end{tabular}
\end{table}

%% file: tables/ablation.tex
\begin{table*}[htbp]
    \centering
    \setlength{\tabcolsep}{4pt}
    \renewcommand{\arraystretch}{0.9}
    \small
    \caption{Ablation study on the design choices of our method using Gemma-2-2B-IT. ASR is reported on TwinBreak; over-refusal on XSTest; utility on two representative tasks.}
    \label{tab:ablation}
    

                

    \begin{tabular}{ll cccc}
        \toprule
        & \multirow{2}{*}{\textbf{Setting}} & \textbf{ASR} $\downarrow$ & \textbf{Over-refusal} $\downarrow$ & \multicolumn{2}{c}{\textbf{Utility Difference}} \\
        \cmidrule(lr){5-6}
         & & \small TwinBreak~\cite{twinbreak} & \small XSTest~\cite{rottger2024xstest} & Arc Challenge~\cite{arc_challenge} & WinoGrande~\cite{WinoGrande} \\
        \midrule
        1 & No Pruning                                          & 0.52 & 7.0\% & +0.09\% & -0.39\% \\
        2 & Prune Random Neurons instead of Safety Neurons      & 0.50 & 6.0\% & +0.09\% & -0.24\% \\
        \midrule
        3 & Non-gradual pruning                                 & 0.38 & 3.0\% & +0.43\% & -0.63\% \\
        4 & Decremental pruning                                 & 0.31 & 3.0\% & +0.09\% & -1.50\% \\
        5 & Prune Randomly Sampled Safety Neurons ($r=0.2$)     & 0.41 & 2.0\% & +0.26\% & -0.32\% \\
          & \hspace{160pt}($r=0.4$)                             & 0.32 & 4.0\% & -0.17\% & -1.34\% \\
          & \hspace{160pt}($r=0.6$)                             & 0.31 & 4.0\% & -0.09\% & -1.42\% \\

        \midrule
        6 & Unfreeze First and Last Transformer Layers          & 0.23 & 7.0\% & -1.09\% & -1.95\% \\
        7 & Longer Train  (4 Epochs)                            & 0.32 & 5.0\% & -0.43\% & -1.74\% \\
        8 & Shorter Train (1 Epoch)                             & 0.45 & 3.0\% & -0.09\% & -0.71\% \\
                
        \midrule
        - & Ours (Full) & 0.23 & 4.0\% & 0.00\% & -1.50\% \\
        \bottomrule
    \end{tabular}
\end{table*}

%% file: tables/adaptive.tex
\begin{table}[htbp]
    \centering
    \setlength{\tabcolsep}{4pt}
    \renewcommand{\arraystretch}{0.9}
    \small
    \caption{Attack success rate (ASR) against adaptive attackers with increasing iteration budgets, comparing no defense vs.\ our method.}
    \label{tab:adaptive_attack}
    
        
        

    \begin{tabular}{l cccc}
        \toprule
        \textbf{Model} & \textbf{ASR@5} & \textbf{ASR@10} & \textbf{ASR@15} & \textbf{No Def.} \\
        \midrule 
        
        Gemma-2-2B-IT           & 0.23 & 0.26 & 0.27  & 0.71 \\
        Llama-3-8B-Instruct     & 0.11 & 0.25  & 0.25 & 0.81  \\
        Llama-3.2-3B-Instruct   & 0.33 & 0.38 & 0.39  & 0.74 \\
        Qwen-3-4B               & 0.13 & 0.22 & 0.29  & 0.79 \\
        
        \bottomrule
    \end{tabular}
\end{table}

%% file: text/05_discusssion.tex
\section{Discussion}
\label{sec:discussion}

In this section, we discuss the limitations of our work, along with insights into unresolved challenges and directions for future work that emerged from our findings.

\subsection{Limitations}
\label{subsec:limitations}

\textbf{Safety Neuron Detection Methods Are Not Perfect.} \method relies on safety neuron detection methods~\cite{twinbreak, wei2023jailbroken, wu2026neurostrike}. These methods typically rely on heuristics, such as identifying the top $k$ neurons with the largest activation differences or highest scores. As a result, they may not identify a perfect set of safety-critical neurons for the target LLM: some safety neurons may be missed, while a small number of unrelated neurons may be incorrectly identified as safety-critical. Furthermore, some neurons have been shown to serve multiple functions~\cite{wei2023jailbroken}, making it difficult to associate individual neurons exclusively with safety alignment.

The imperfection of safety neuron detection has two important implications for \method. First, since some safety-critical neurons may remain undetected, it is difficult to obtain a fully unaligned model without also degrading its general capabilities. As a result, the model may continue to rely on parts of the existing safety mechanism during finetuning, making it more difficult to induce completely new and more distributed safety representations. Second, pruning incorrectly identified or multifunctional neurons may lead to unintended behavior or degradation in general model capabilities, either because these neurons serve multiple functions or because they are not actually part of the safety mechanism.

\textbf{Measuring Safety-Alignment Density.} Our results show that ABS-Align reduces the model’s reliance on the safety-critical neurons identified in the original model, as evidenced by its increased robustness to multiple attacks that exploit sparse safety representations. However, given the limitations of existing safety-neuron detection methods, we do not directly quantify the absolute density of safety alignment in the resulting model. We therefore use denser safety alignment to refer to the less fragile and reduced dependence of safety-aligned behavior on the originally identified sparse set of safety-critical neurons.

\textbf{Overhead of Preference Optimization.} \method relies on DPO~\cite{rafailov2023dpo} for preference-based finetuning. While DPO is substantially more computationally efficient than RLHF-based alignment methods~\cite{rafailov2023dpo, ouyang2022rlhf}, it still incurs considerable computational and memory overhead because it requires maintaining a frozen reference model in addition to the model being finetuned.

\subsection{Future Work}
\label{subsec:future_work}

Safety alignment is not the only model behavior that has been shown to be tied to a small number of neurons in the model architecture; recent works have reported similar phenomena for various tasks and languages~\cite{tang2024language, wang2022findingskill}. A potential direction for future research is to investigate whether the proposed method can be extended to make representations of specific capabilities more distributed across the model, with potential applications in settings where a model underperforms on a particular task or where improving performance on a specific task is desired.

Another interesting direction for future work is to better understand the post-training process, particularly safety alignment, and investigate whether non-sparse refusal representations can be established during the main stage of preference optimization rather than through post-hoc approaches such as \method.

%% file: text/06_related_work.tex
\section{Related Work}
\label{sec:related_work}
This section provides an overview of related work in three main categories. First, Section~\ref{subsec:related_work:jailbreak_attacks} reviews representative jailbreak methods, including both black-box and white-box attacks. Next, Section~\ref{subsec:related_work:defenses} discusses defensive approaches proposed to mitigate jailbreak attacks. Finally, Section~\ref{subsec:related_work:safety_alignment} reviews attacks that exploit the sparsity of safety alignment in LLMs.

\subsection{Jailbreak Attacks}
\label{subsec:related_work:jailbreak_attacks}
A broad range of works have shown that both open-weight and closed-source LLMs are vulnerable to jailbreak attacks, where attackers use various techniques to elicit outputs that the models would otherwise refuse to generate~\cite{wei2023jailbroken, chao2025pair, andriushchenko2025simpleadaptive, twinbreak, wu2026neurostrike, arditi2024refusal, ding2023renellm}. These attacks can be broadly categorized into black-box and white-box approaches based on the level of access the attacker has to the target model.

In black-box settings, the attacker only has query access to the target LLM wherein the attacker can only prompt the target model and observer the outputs returned by the LLM. Black-box attacks initially often relied on manually crafted adversarial prompts employing techniques such as role-playing, persuasion, or obfuscation of malicious intent~\cite{johnny2024, dualuse, dan2024}. Later works tried to make this more systematic~\cite{chao2025pair, ding2023renellm, mehrotra2024tap}. PAIR~\cite{chao2025pair, llmfuzzer} for instance uses an attacker LLM to iteratively refine and improve jailbreak prompts using only black-box access to the target LLM where the attacker LLM creates adversarial prompts, feeds it to the target LLM and refines its output based on the target LLMs response. 

In white-box scenario, the attacker has access to the target LLM's architecture, weights, and internal states~\cite{zou2023GCG, liu2024autodan, wu2026neurostrike, twinbreak, arditi2024refusal}. This additional access enables attacks that are unavailable in black-box settings, including computing model gradients~\cite{zou2023GCG}, directly intervening on model parameters~\cite{wu2026neurostrike, twinbreak}, and manipulating intermediate activations~\cite{arditi2024refusal} during inference. GCG~\cite{zou2023GCG}, for instance, appends a set of adversarial suffix tokens to an input prompt and uses gradients from the open-weight target LLM to optimize the suffix tokens toward eliciting an affirmative response to a harmful request. Another example is AutoDAN~\cite{liu2024autodan}, which uses a hierarchical genetic algorithm to iteratively improve a population of jailbreak prompts based on the target model's likelihood of generating a predefined affirmative response.

\subsection{Defense Methods against Jailbreaks}
\label{subsec:related_work:defenses}

Improving the robustness of LLMs against jailbreaks has emerged as a challenging problem and has resulted in a growing body of work on defensive approaches. At a high-level, existing defenses against jailbreaks can be broadly categorized into input-time defenses, output-time defenses, and finetuning-based methods. 

\subsubsection{Input-time and Output-time defenses}

Input-time defenses typically rely on tools that are external to the LLM itself and process the input before it is fed to the LLM~\cite{alon2023perplexity, kumar2024certifying, Inan2023llamaguard}. Alon et al.~\cite{alon2023perplexity} proposed using perplexity (a metric measuring how likely the LLM is to generate a particular token sequence) to detect adversarial jailbreak prompts, motivated by the fact that early jailbreak prompts commonly employed adversarial suffixes to bypass safety filters~\cite{zou2023GCG}. Kumar et al.~\cite{kumar2024certifying} proposed an "erase-and-check" approach that generates subsequences by erasing tokens from the input prompt and inspects them using a safety filter. The input is flagged as harmful if either the original prompt or any of the inspected subsequences is classified as harmful. Llama Guard~\cite{Inan2023llamaguard} was proposed by Inan et al. as an LLM-based safeguard model that scans user inputs (and additionally model outputs) to determine whether it contains unsafe content according to a safety-risk taxonomy. SmoothLLM~\cite{robey2025smoothllm} defends against jailbreak attacks by creating multiple randomly perturbed copies of an input prompt and aggregating model responses to each perturbed version based on the intuition that adversarial prompts are fragile to small scale modifications. A majority vote is then conducted on the aggregated responses and a response consistent with the majority vote is returned by the model.

One of the disadvantages of input-time defenses is the additional overhead they introduce by filtering or transforming the inputs before forwarding user requests to the deployed LLM. Furthermore, these defenses don't improve the inherent robustness of the model; therefore, if the defense fails or is circumvented, the underlying LLM remains vulnerable to jailbreak attacks.

Output-time defenses are also deployed externally and process responses generated by the LLM before returning them back to the user~\cite{Inan2023llamaguard, ji2024aligner}. Aligner~\cite{ji2024aligner}, for instance, proposes an approach that processes output responses and applies learned corrections to the output text to reduce harmful content and improve overall alignment. Similar to input-time defenses, however, output-time defenses do not modify the underlying safety alignment of the target LLM and therefore retain many of the same limitations.

Unlike output-time and input-time defenses, \method directly finetunes the model therefore improving its inherent safety robustness instead of relying on external mechanisms that can only process inputs or outputs of models.

\subsubsection{Finetuning-based Defenses}

Fine-tuning-based defenses directly finetune an LLM to improve its robustness against jailbreak attacks~\cite{zou2024circuit, hu2026alignmentweighted, qi2025deep, zhao2025door}. W-DOOR~\cite{zhao2025door} improves safety alignment by decomposing DPO's loss into a dual objective loss that optimizes refusal training in addition to unlearning to targeted unlearning of harmful knowledge, improving robustness against a range of jailbreak attacks such as prefilling~\cite{wei2023jailbroken} and adversarial suffix attacks~\cite{zou2023GCG}.

Zou et al.~\cite{zou2024circuit} propose using representation finetuning in order to "circuit break" LLMs by rerouting harmful internal representations away from trajectories that would otherwise lead to harmful completions. In other words, they train LLMs into redirecting harmful representation states as soon as they emerge in the model's internals while preserving benign representations, essentially "short-circuiting" the internal trajectories that lead to harmful outputs. 

Hu et al. propose Alignment-Weighted DPO~\cite{hu2026alignmentweighted} which uses a Chain-of-Thought safety alignment dataset in addition to a fine-grained training method that decomposes model responses in the training dataset into two parts, namely the reasoning trace and final answer. The method assigns different weights to these two parts of the response based on the strength of the correction signal required for each part.

None of these works address the sparse and fragile safety alignment of LLMs leaving them vulnerable to intervention-based attacks~\cite{arditi2024refusal} and neuron-pruning attacks~\cite{wu2026neurostrike, twinbreak}. In contrast, \method directly targets the sparsity of LLM safety alignment by finetuning the model while progressively pruning its safety-critical neurons, encouraging safety-aligned behavior to become distributed across other parts of the model.

\subsection{Sparsity of Safety Alignment}
\label{subsec:related_work:safety_alignment}

In recent years, multiple works have shown that safety alignment in modern large language models is highly fragile due to the sparse nature of the resulting safety mechanisms in safety-aligned LLMs~\cite{arditi2024refusal, twinbreak,wu2026neurostrike,wei2024brittleness}. In other words, prior work has shown that safety mechanisms in LLMs have a low density both in the model's internal representation space~\cite{arditi2024refusal} and in their distribution across the model architecture~\cite{twinbreak, wu2026neurostrike, wei2024brittleness}.

Architecture-wise sparsity in safety alignment allows attackers to jailbreak LLMs by finding and removing safety-critical neurons, essentially removing the safety alignment and transforming the pruned model into a jailbroken version. Numerous works have shown that pruning only 1-5\% of neurons can substantially degrade safety alignment while having little impact on the model's general capabilities~\cite{twinbreak, wei2024brittleness, wu2026neurostrike}.

Wei et al.~\cite{wei2024brittleness} use Wanda~\cite{sun2024wanda} and SNIP~\cite{lee2018snip} scores computed on separate benign and harmful prompts to find crucial neurons for safety alignment and removing them. TwinBreak~\cite{twinbreak} inputs structurally and semantically similar pairs of benign and malicious prompts to a target LLM and compares the resulting intermediate activations between the two sets to locate and prune neurons in an iterative manner. NeuroStrike~\cite{wu2026neurostrike} similarly identifies and prunes safety neurons by training linear probes on intermediate activations from benign and malicious prompts to identify neurons that are strongly associated with distinguishing harmful inputs. Furthermore, Arditi et al.~\cite{arditi2024refusal} show that refusal behavior in language models is represented by a single direction in the model's activation space and that by ablating this direction from the activation space it is possible to jailbreak the target LLM.

One of the main goals of \method is to directly address this weakness. Our method reinforces safety alignment by gradually pruning safety-critical neurons and re-establishing safety alignment by finetuning the resulting pruned model. This process encourages safety-aligned behavior to become distributed across a broader set of neurons and representations, resulting in a less fragile and more robust safety-aligned model.

%% file: text/07_conclusion.tex
\section{Conclusion}
\label{sec:conclusion}


We propose \method, a method for making safety alignment more resilient and less fragile, and show that this leads to broad improvements across multiple dimensions compared to existing defenses while requiring only limited finetuning. \method significantly reduces the effectiveness of attacks that exploit sparse neuron-level or representation-level safety alignment, while also improving robustness against a wide range of other jailbreak attacks without any explicit guidance or data augmentation during finetuning.

%% file: text/99_appendix.tex
\appendix

\renewcommand{\thesubsection}{\arabic{subsection}}

\subsection{Training Details}
\label{appendix:training}
We train all the evaluated models for two epochs on 1,000 randomly selected sample from PKU-SafeRLHF~\cite{ji2024pku} dataset with a batch size of two. Following~\cite{zou2024circuit}, we observed that adding a small number of samples for retaining models' general capability is necessary. Therefore we follow Circuit Breaker and additionally use 850 samples from Alpaca~\cite{alpaca} and 150 benign samples from XSTest~\cite{rottger2024xstest} during training Llama-3-8B-Instruct and Llama-3.2-3B to retain the models' capability during fine-tuning. To create preference pairs for the retain part of the dataset, we use a set of fixed refusal answers for the dispreferred choice and each model's benign response for the preferred choice. Preference pairs for PKU-SafeRLHF are used directly from the dataset itself.

Regarding ABS-Align's pruning strategy, we set pruning steps $t_p = 200$ and the pruning rate to 20\% such that every 200 steps safety neurons corresponding to one pruning iteration detected by TwinBreak are pruned. We use a linear cosine scheduler and AdamW as the optimizer for both models.

For finetuning the Gemma-2-2B-IT model, we use a learning rate of $1e-5$ and we set the beta for DPO to 0.1. We set weight decay to 0 and warmup ratio to 0.1. For Finetuning Llama-3-8B-Instruct, we use a learning rate of 2e-5, beta of 0.1, and warmup ratio of 0.1. For Llama-3.2-3B we use a learning rate of 2e-4 and beta of 0.3 with a warmup ratio of 0.1. Finally, for Qwen-3-4B we use a learning rate of 2e-4, a beta of 0.1, and a warmup ratio of 0.05.

\subsection{SORRY-Bench Black-box-style Adversarial Prompts}
\label{appendix:sorry_bench}
SORRY-Bench~\cite{xie2025sorrybench} has black-box twenty attack types in four categories namely writing style attacks, multi-languages, encoding and decryption, and presuasion techniques. We exclude multi-language and encoding and decryption attacks and select the ten categories in Table~\ref{tab:sorrybench_attacks} in our evaluations.

\begin{table}[t]
    \centering
    \small
    \setlength{\tabcolsep}{4pt}
    \renewcommand{\arraystretch}{0.9}
    \caption{SORRY-Bench black-box attack transformations used in our evaluation.}
    \label{tab:sorrybench_attacks}
    \begin{tabular}{p{0.28\linewidth} p{0.64\linewidth}}
        \toprule
        \textbf{Attack Type} & \textbf{Description} \\
        \midrule
        Question & Harmful instruction rephrased as question \\

        Slang & Rewrites the harmful prompt using colloquial language and slang. \\

        Uncommon Dialects & Expresses the harmful request using less common dialects. \\

        Technical Terms & Rephrases the request using technical terminology. \\

        Role Play & Frames the harmful request within an assigned role or hypothetical persona. \\

        Misspellings & Introduces spelling or grammatical errors into the request. \\

        Authority Endorsement & Cites authoritative sources or institutions to support the request \\

        Misrepresentation & Misrepresents the harmful purpose of the request as benign. \\

        Evidence-based Persuasion & Uses facts to justify fulfilling the request. \\

        Expert Endorsement & Invokes the opinion of an expert. \\
        \bottomrule
    \end{tabular}
\end{table}

\subsection{Full SORRY-Bench Results}
\label{appendix:full_sorry_bench}
We presented ASR over five attacks from SORRY-Bench and the average over all the selected ten attacks in Section~\ref{subsubsec:robustness_black_box}. Here we provide the full results over all the ten attacks for completeness. Table~\ref{tab:sorrybench_full_gemma} provides ASR for all the ten evaluated attacks for Gemma-2-2B-IT which shows that ABS-Align consistently improves robustness. Results for Llama-3-8B-Instruct, Llama-3.2-3B, and Qwen-3-4B show a similar trend as the numbers reported in Tables~\ref{tab:sorrybench_full_llama}, ~\ref{tab:sorrybench_full_llama3.2}, and ~\ref{tab:sorrybench_full_qwen} show.

\begin{table}[t]
    \centering
    \setlength{\tabcolsep}{6pt}
    \renewcommand{\arraystretch}{0.95}
    \small
    \caption{Attack success rate (\%) of Gemma-2-2B-IT across all ten SORRY-Bench black-box attacks used in our evaluation. Lower is better.}
    \label{tab:sorrybench_full_gemma}
    \begin{tabular}{lcc}
        \toprule
        \textbf{Attack} & \textbf{No Defense} & \textbf{ABS-Align} \\
        \midrule
        Question                    & 19.77 & \textbf{12.50} \\
        Slang                       & 31.59 & \textbf{13.86} \\
        Uncommon Dialects           & 27.95 & \textbf{12.50} \\
        Technical Terms             & 24.09 & \textbf{8.86} \\
        Role Play                   & 20.68 & \textbf{6.81} \\
        Misspellings                & 30.90 & \textbf{12.04} \\
        Authority Endorsement       & 65.45 & \textbf{20.45} \\
        Misrepresentation           & 72.95 & \textbf{17.27} \\
        Evidence-based Persuasion   & 55.68 & \textbf{11.81} \\
        Expert Endorsement          & 66.81 & \textbf{16.36} \\
        \midrule
        \textbf{Average}            & \textbf{41.58} & \textbf{13.24} \\
        \bottomrule
    \end{tabular}
\end{table}

\begin{table}[t]
    \centering
    \setlength{\tabcolsep}{4pt}
    \renewcommand{\arraystretch}{0.9}
    \small
    \caption{Attack success rate (\%) of Llama-3-8B-Instruct across all ten SORRY-Bench black-box attacks used in our evaluation. Lower is better.}
    \label{tab:sorrybench_full_llama}
    \begin{tabular}{lcc}
        \toprule
        \textbf{Attack} & \textbf{No Defense} & \textbf{ABS-Align} \\
        \midrule
        Question                    & 41.13 & \textbf{18.86} \\
        Slang                       & 41.36 & \textbf{16.59} \\
        Uncommon Dialects           & 41.36 & \textbf{11.36} \\
        Technical Terms             & 46.36 & \textbf{20.00} \\
        Role Play                   & 36.36 & \textbf{15.90} \\
        Misspellings                & 43.18 & \textbf{18.40} \\
        Authority Endorsement       & 60.90 & \textbf{22.50} \\
        Misrepresentation           & 55.45 & \textbf{13.63} \\
        Evidence-based Persuasion   & 56.59 & \textbf{22.04} \\
        Expert Endorsement          & 60.68 & \textbf{18.63} \\
        \midrule
        \textbf{Average}            & \textbf{48.34} & \textbf{17.79} \\
        \bottomrule
    \end{tabular}
\end{table}

\begin{table}[t]
    \centering
    \setlength{\tabcolsep}{4pt}
    \renewcommand{\arraystretch}{0.9}
    \small
    \caption{Attack success rate (\%) of Llama-3.2-3B across all ten SORRY-Bench black-box attacks used in our evaluation. Lower is better.}
    \label{tab:sorrybench_full_llama3.2}
    \begin{tabular}{lcc}
    \toprule
    \textbf{Attack} & \textbf{No Defense} & \textbf{ABS-Align} \\
    \midrule
    Question                    & 36.82 & \textbf{27.27} \\
    Slang                       & 36.36 & \textbf{22.73} \\
    Uncommon Dialects           & 35.23 & \textbf{20.91} \\
    Technical Terms             & 43.18 & \textbf{27.95} \\
    Role Play                   & 34.55 & \textbf{22.05} \\
    Misspellings                & 34.77 & \textbf{20.00} \\
    Authority Endorsement       & 79.77 & \textbf{49.09} \\
    Misrepresentation           & 83.41 & \textbf{48.18} \\
    Evidence-based Persuasion   & 71.14 & \textbf{46.36} \\
    Expert Endorsement          & 75.45 & \textbf{45.68} \\
    \midrule
    \textbf{Average}            & \textbf{53.07} & \textbf{33.02} \\
    \bottomrule
\end{tabular}
\end{table}

\begin{table}[t]
    \centering
    \setlength{\tabcolsep}{4pt}
    \renewcommand{\arraystretch}{0.9}
    \small
    \caption{Attack success rate (\%) of Qwen-3-4B across all ten SORRY-Bench black-box attacks used in our evaluation. Lower is better.}
    \label{tab:sorrybench_full_qwen}
    \begin{tabular}{lcc}
        \toprule
        \textbf{Attack} & \textbf{No Defense} & \textbf{ABS-Align} \\
        \midrule
        Question                    & 44.09 & \textbf{6.81} \\
        Slang                       & 55.90 & \textbf{9.54} \\
        Uncommon Dialects           & 58.18 & \textbf{13.86} \\
        Technical Terms             & 52.50 & \textbf{13.40} \\
        Role Play                   & 52.04 & \textbf{7.04} \\
        Misspellings                & 50.45 & \textbf{8.40} \\
        Authority Endorsement       & 93.63 & \textbf{49.09} \\
        Misrepresentation           & 92.95 & \textbf{58.18} \\
        Evidence-based Persuasion   & 93.86 & \textbf{47.72} \\
        Expert Endorsement          & 93.86 & \textbf{57.95} \\
        \midrule
        \textbf{Average}            & \textbf{68.75} & \textbf{27.20} \\
        \bottomrule
    \end{tabular}
\end{table}